\documentclass[10pt,journal,cspaper,compsoc]{IEEEtran}

\usepackage{times}
\usepackage{epsfig}
\usepackage{graphicx}
\usepackage[fleqn]{amsmath}
\usepackage{amssymb}
\usepackage{booktabs}
\usepackage{multirow}
\usepackage{cite}
\usepackage{hyperref}
\usepackage{url}

\begin{document}

\title{Places: An Image Database for Deep Scene Understanding}

\author{Bolei~Zhou, ~Aditya~Khosla, ~Agata~Lapedriza,~Antonio~Torralba
        and~Aude~Oliva% <-this % stops a space
\IEEEcompsocitemizethanks{
\IEEEcompsocthanksitem B. Zhou, A. Khosla, A.Torralba, A.Oliva are with the Computer Science and Artificial Intelligence Laboratory, Massachusetts Institute of Technology, USA. 
\IEEEcompsocthanksitem A. Lapedriza is with Universitat Oberta de Catalunya, Spain.
}
}

%\markboth{IEEE TRANSACTIONS ON PATTERN ANALYSIS AND MACHINE INTELLIGENCE, VOL.XX, No.XX, XX}%
%{Shell \MakeLowercase{\textit{et al.}}: Bare Demo of IEEEtran.cls for Computer Society Journals}

\IEEEcompsoctitleabstractindextext{%

\begin{abstract}

The rise of multi-million-item dataset initiatives has enabled data-hungry machine learning algorithms to reach near-human semantic classification at tasks such as object and scene recognition. Here we describe the Places Database, a repository of 10 million scene photographs, labeled with scene semantic categories and attributes, comprising a quasi-exhaustive list of the types of environments encountered in the world. Using state of the art Convolutional Neural Networks, we provide impressive baseline performances at scene classification. With its high-coverage and high-diversity of exemplars, the Places Database offers an ecosystem to guide future progress on currently intractable visual recognition problems.  

\end{abstract}

}

\maketitle

\IEEEdisplaynotcompsoctitleabstractindextext

\section{Introduction}

What does it take to reach human-level performance with a machine-learning algorithm? In the case of supervised learning, the problem is two-fold. First, the algorithm must be suitable for the task, such as pattern classification in the case of object recognition \cite{krizhevsky2012imagenet,zhou2014learning}, pattern localization for object detection \cite{girshick2014rich} or the necessity of temporal connections between different memory units for natural language processing \cite{hochreiter1997long,manning1999foundations}. Second, it must have access to a training dataset of appropriate coverage (quasi-exhaustive representation of classes and variety of examplars) and density (enough samples to cover the diversity of each class). The optimal space for these datasets is often task-dependent, but the rise of multi-million-item sets has enabled unprecedented performance in many domains of artificial intelligence.

The successes of Deep Blue in chess, Watson in ``Jeopardy!", and AlphaGo in Go against their expert human opponents may thus be seen as not just advances in algorithms, but the increasing availability of very large datasets: 700,000, 8.6 million, and 30 million items, respectively \cite{campbell2002deep,ferrucci2013watson,silver2016mastering}. Convolutional Neural Networks \cite{krizhevsky2012imagenet,lecun1998gradient} have likewise achieved near human-level visual recognition, trained on 1.2 million object \cite{he2015delving,deng2009imagenet,russakovsky2015imagenet} and 2.5 million scene images \cite{zhou2014learning}. Expansive coverage of the space of classes and samples allows getting closer to the right ecosystem of data that a natural system, like a human, would experience.

\begin{figure*}
\begin{center}
\includegraphics[width=1\linewidth]{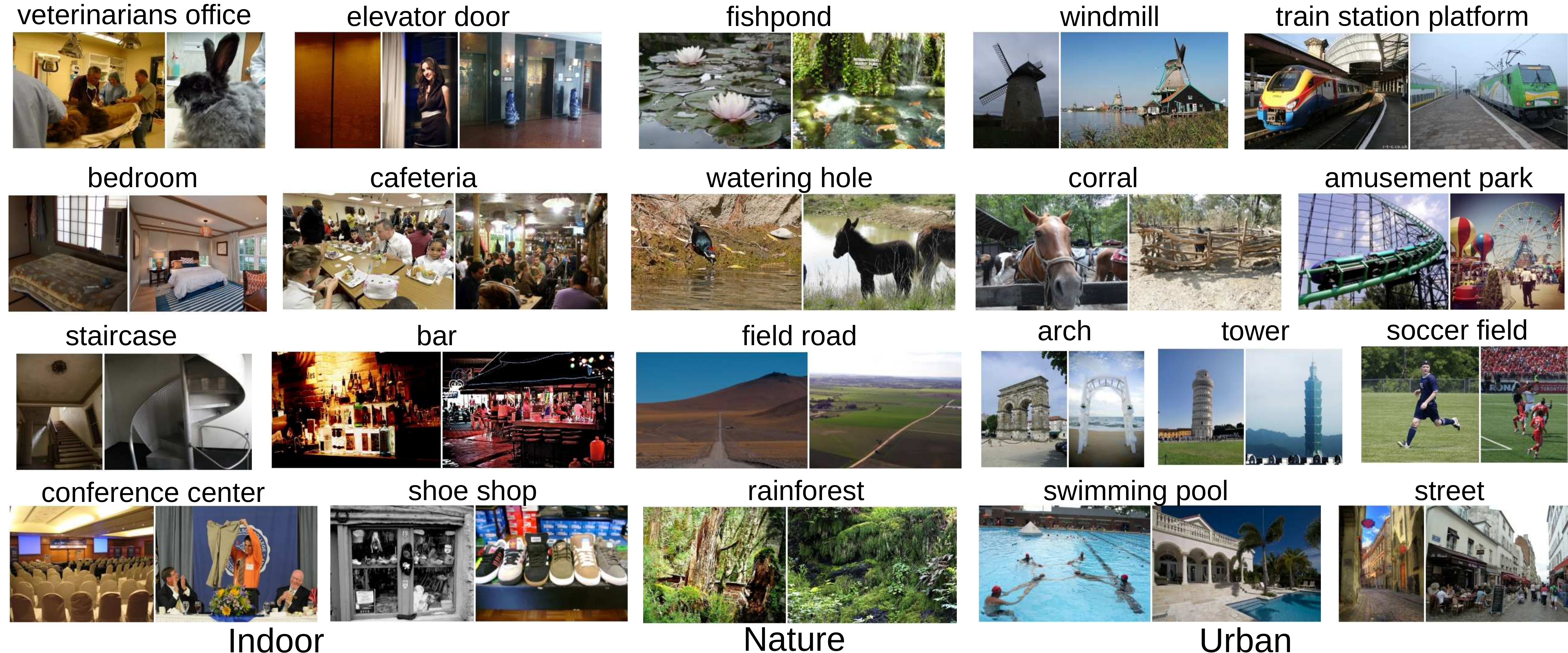}
\end{center}
\caption{Image samples from various categories of the Places Database. The dataset contains three macro-classes: Indoor, Nature, and Urban.}
\label{overview_sample}
\end{figure*}

Here we describe the Places Database, a quasi-exhaustive repository of 10 million scene photographs, labeled with 476 scene semantic categories and attributes, comprising the types of visual environments encountered in the world. Image samples are shown in Fig. \ref{overview_sample}. In the context of Places, we explain the steps to create high-quality datasets enabling the remarkable feats of machine-learning algorithms.

\begin{figure*}
\begin{center}
\includegraphics[width=1\linewidth]{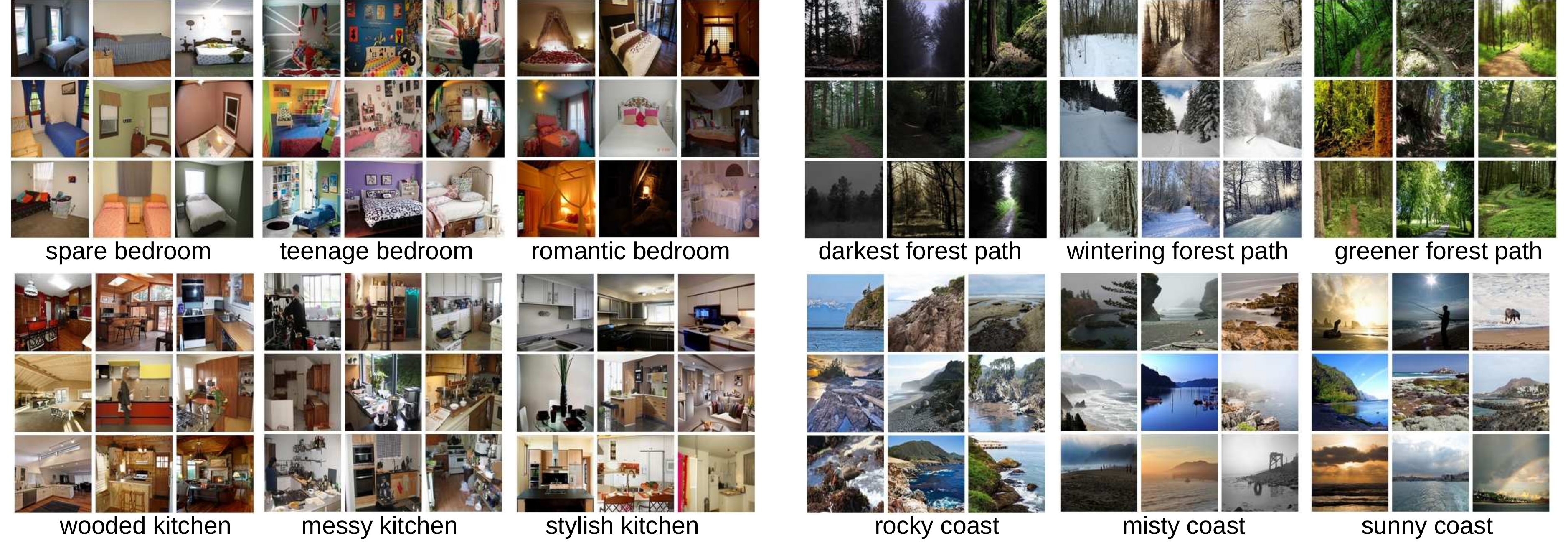}
\end{center}
\caption{Image samples from four scene categories grouped by queries to illustrate the diversity of the dataset. For each query we show 9 annotated images.}
\label{aj_plot}
\end{figure*}

\section{Places Database}

\subsection{Coverage of the categorical space}

The primary asset of a high-quality dataset is an expansive coverage of the categorical space we want to learn. The strategy of Places is to provide an exhaustive list of the categories of environments encountered in the world, bounded by spaces where a human body would fit (e.g. closet, shower). The SUN (Scene UNderstanding) dataset \cite{xiao2010sun} provided that initial list of semantic categories. The SUN dataset was built around a quasi-exhaustive list of scene categories with different functionalities, namely categories with unique identities in discourse. Through the use of WordNet \cite{miller1995wordnet}, the SUN database team selected 70,000 words and concrete terms that described scenes, places and environments that can be used to complete the phrase ``I am in a \textit{place}", or ``let's go to the/a \textit{place}". Most of the words referred to basic and entry-level names (\cite{jolicoeur1984pictures}), resulting in a corpus of 900 different scene categories after bundling together synonyms, and separating classes described by the same word but referring to different environments (e.g. inside and outside views of churches). Details about the building of that initial corpus can be found in \cite{xiao2010sun}. Places Database has inherited the same list of scene categories from the SUN dataset.

\subsection{Construction of the database}

\subsubsection{Step 1: Downloading images using scene category and adjectives}

From online image search engines (Google Images, Bing Images, and Flickr), candidate images were downloaded using a query word from the list of scene classes provided by the SUN database \cite{xiao2010sun}. In order to increase the diversity of visual appearances in the Places dataset (see Fig. \ref{aj_plot}), each scene class query was combined with 696 common English adjectives (e.g., messy, spare, sunny, desolate, etc.). About 60 million images (color images of at least 200$\times$200 pixels size) with unique URLs were identified. Importantly, the Places and SUN datasets are complementary: PCA-based duplicate removal was conducted within each scene category in both databases so that they do not contain the same images.

\subsubsection{Step 2: Labeling images with ground truth category}

\begin{figure*}
\begin{center}
\includegraphics[width=1\linewidth]{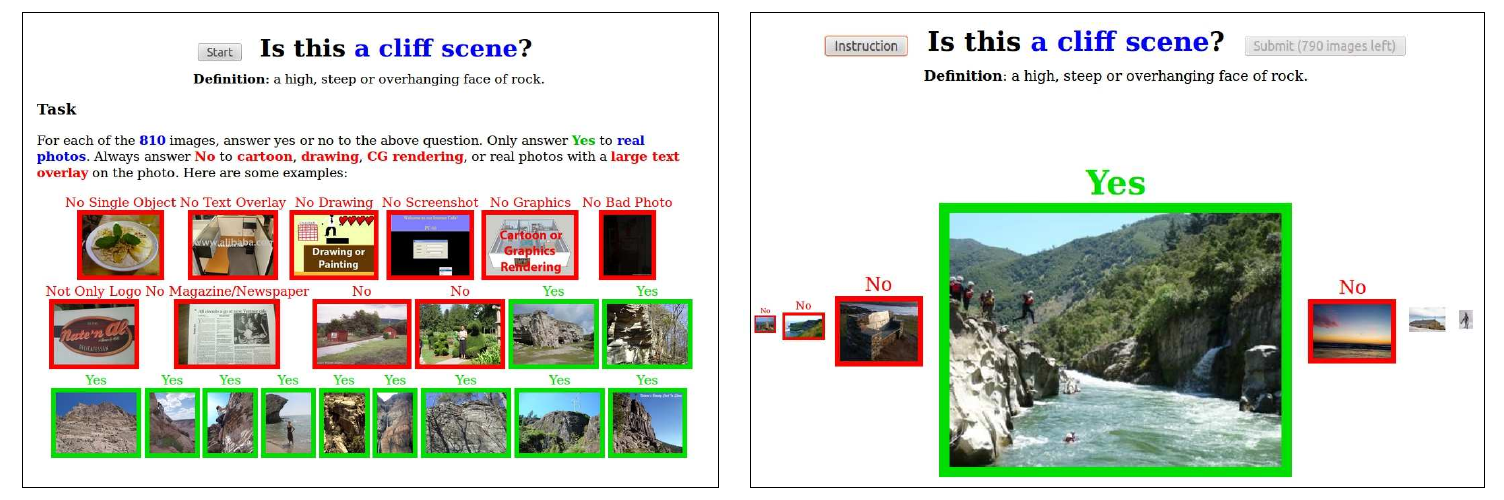}
\end{center}
\caption{Annotation interface in the Amazon Mechanical Turk for selecting the correct exemplars of the scene from the downloaded images. The left plot shows the instruction given to the workers in which we define positive and negative examples. The right plot shows the binary selection interface.}\label{fig:AMT_1}
\end{figure*}

Image ground truth label verification was done by crowdsourcing the task to Amazon Mechanical Turk (AMT). Fig.\ref{fig:AMT_1} illustrates the experimental paradigm used: AMT workers were each given instructions relating to a particular image category at a time (e.g. cliff), with a definition and samples of true and false images.  Workers then performed a go/no-go categorical task (Fig.\ref{fig:AMT_1}). The experimental interface displayed a central image, flanked by smaller version of images the worker had just responded to, on the left, and will respond to next, on the right.  Information gleaned from construction of the SUN dataset suggests that the first iteration of labeling will show that more than 50\% of the the downloaded images are not true exemplars of the category.  As illustrated in Fig.\ref{fig:AMT_1}, the default answer is set to \textit{No} (see images with bold red contours), so the worker can more easily press the space bar to move the majority of No images forward. Whenever a true category exemplar appears in the center, the worker can press a specific key to mark it as a positive exemplar (responding \textit{yes} to the question: ``is this a \textit{place term}"). Reaction time from the moment the image is centrally placed to the space bar or key press is recorded. The interface also allows moving backwards to revise previous annotations. Each AMT HIT (Human Intelligence Task, one assignment for one worker), consisted of 750 images for manual annotation. A control set of 30 positive samples and 30 negative samples with ground-truth category labels from the SUN database were intermixed in the HIT as well. Only worker HITs with an accuracy of 90\% or higher on these control images were kept. 

The positive images resulting from the first cleaning iteration were sent for a second iteration of cleaning. We used the same task interface but with the default answer set to \textit{Yes}. In this second iteration, 25.4\% of the images were relabeled as \textit{No}. We tested a third iteration on a few exemplars but did not pursue it further as the percentage of images relabeled as No was not significant.

After the two iterations of annotation, we collected one scene label for 7,076,580 images pertaining to 476 scene categories. As expected, the number of images per scene category vary greatly (i.e. there are many more images of bedroom than cave on the web). There were 413 scene categories that ended up with at least 1000 exemplars, and 98 scene categories with more than 20,000 exemplars. 

\subsubsection{Step 3: Scaling up the dataset using a classifier}

\begin{figure*}
\begin{center}
\includegraphics[width=1\linewidth]{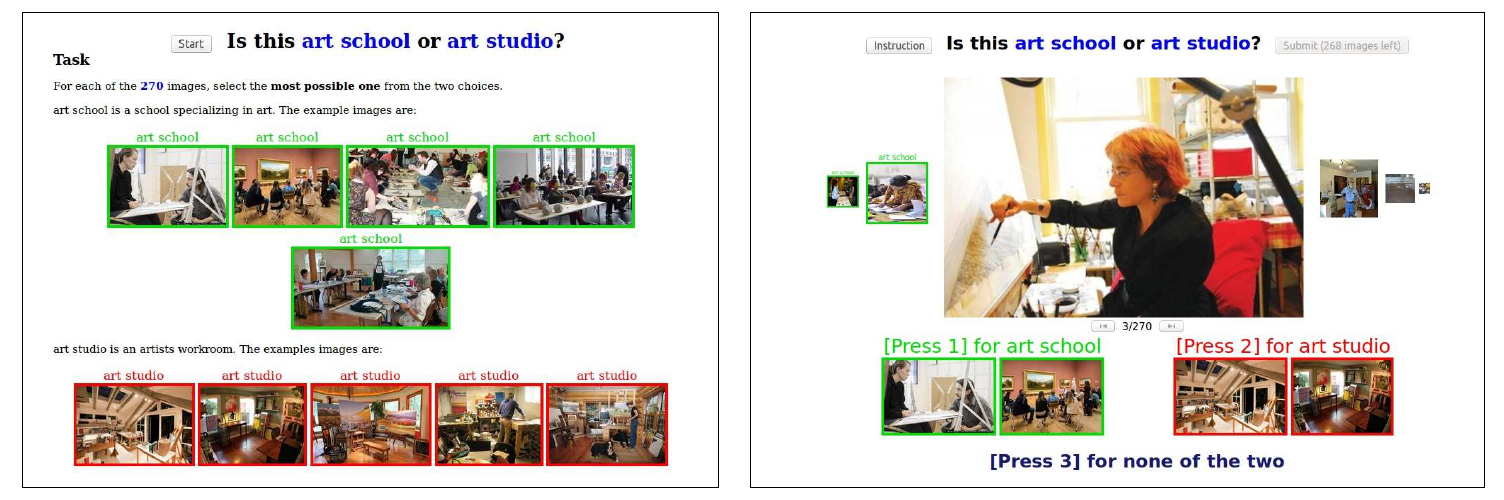}
\end{center}
\caption{Annotation interface in Amazon Mechanical Turk for differentiating images from two similar categories. The left plot shows the instruction in which we give several typical examples in each category. The right plot shows the binary selection interface, in which the worker needs to select the shown image into either of the class or none.}\label{fig:AMT_2}
\end{figure*}

As a result of the previous round of image annotation, there were 53 million remaining downloaded images not assigned to any of the 476 scene categories (e.g. a \textit{bedroom} picture could have been downloaded when querying images for \textit{living-room} category, but marked as negative by the AMT worker). Therefore, a third annotation task was designed to re-classify then re-annotate those images, using a semi-automatic bootstrapping approach. 

A deep learning-based scene classifier, AlexNet \cite{krizhevsky2012imagenet}, was trained to classify the remaining 53 million images: We first randomly selected 1,000 images per scene category as training set and 50 images as validation set (for the 413 categories which had more than 1000 samples).  AlexNet achieved 32\% scene classification accuracy on the validation set after training and was then used to classify the 53 million images. We used the predicted class score by the AlexNet to rank the images within one scene category as follow: for a given category with too few exemplars, the top ranked images with predicted class confidence higher than 0.8 were sent to AMT for a third round of manual annotation using the same interface shown in Fig.\ref{fig:AMT_1}. The default answer was set to \textit{No}.  

After completing the third round of AMT annotation, the distribution of the number of images per category flattened out: 401 scene categories had more than 5,000 images per category and 240 scene categories had more than 20,000 images. Totally there are about 3 million images added into the dataset.

\subsubsection{Step 4: Improving the separation of similar classes}

Despite the initial effort to bundle synonyms from WordNet, the scene list from the SUN database still contained categories with very close synonyms (e.g. `ski lodge' and `ski resort', or `garbage dump' and `landfill'). We identified 46 synonym pairs like these and merged their images into a single category. 

\begin{figure}
\begin{center}
\includegraphics[width=1\linewidth]{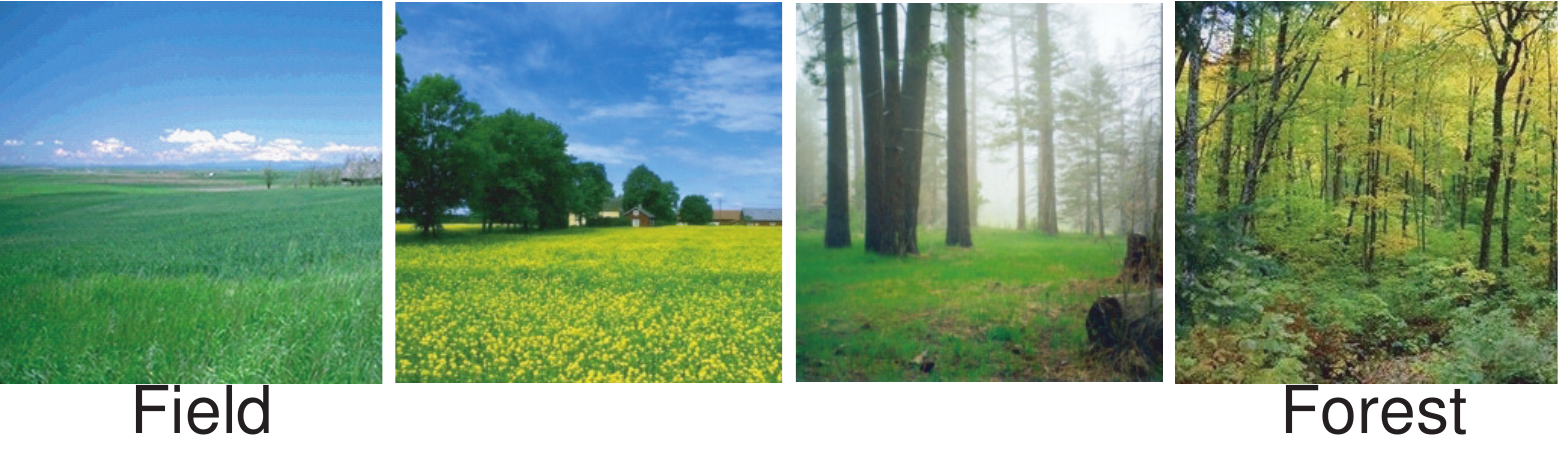}
\end{center}
\caption{Boundaries between place categories can be blurry, as some images can be made of a mixture of different components. The images shown in this figure show a soft transition between a field and a forest. Although the extreme images can be easily classified as field and forest scenes, the middle images can be ambiguous.}
\label{fig:categoryBoundary}
\end{figure}

Additionally, some scene categories are easily confused with blurry categorical boundaries, as illustrated in Fig.~\ref{fig:categoryBoundary}. This means that answering the question ``Does image I belong to class A?" might be difficult. It is easier to answer the question ``Does image I belong to class A or B?" In that case, the decision boundary becomes clearer for a human observer and it also gets closer to the final task that a computer system will be trained to solve.

Indeed, in the previous three steps of the AMT annotation, it became apparent that workers were confused with some pairs of scene categories, for instance, putting images of `canyon' and `butte' into `mountain', or putting `jacuzzi' into `swimming pool indoor', mixing images of `pond' and 'lake', `volcano' and `mountain', `runway' and `landing deck', `highway and road', `operating room' and `hospital room', etc. In the whole set of categories, we identified 53 such ambiguous pairs. 

To further differentiate the images from the categories with shared content, we designed a new interface (Fig.~\ref{fig:AMT_2}) for a fourth step of annotation. We combined exemplar images from the two categories with shared content (such as art school and art studio), and asked the AMT workers to classify images into either of the categories or neither of them. 

After the four steps of annotations, the Places database was finalized with over 10 millions labeled exemplars (10,624,928 images) from 434 place categories. 

\subsection{Scene-Centric Datasets}

\begin{figure*}
\begin{center}
\includegraphics[width=1\linewidth]{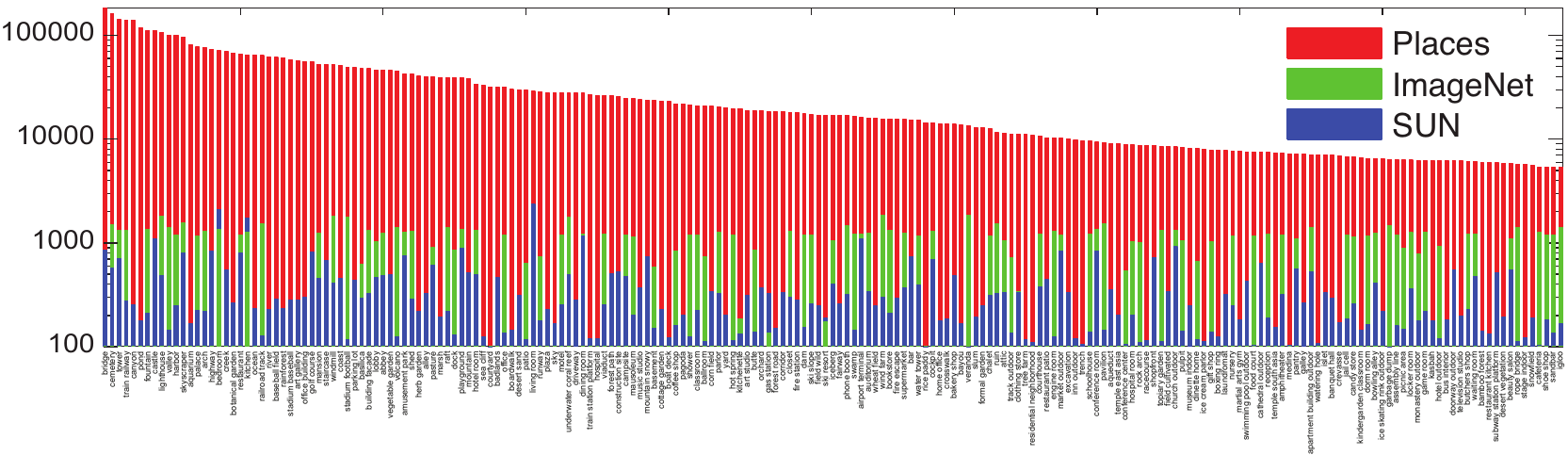}
\end{center}
   \caption{Comparison of the number of images per scene category for the common 88 scene categories in Places, ImageNet, and SUN datasets.}\label{scale}
\end{figure*}

Scene-centric datasets correspond to images labeled with a scene, or place name, as opposed to an object name. Fig.~\ref{scale} illustrates the differences among the number of images found in Places, ImageNet and SUN for a set of scene categories common to all three datasets. Places Database is the largest scene-centric image dataset so far.

\subsubsection{Defining the Benchmarks of the Places}

Here we describe four subsets of Places as benchmarks. Places205 and Places88 are from \cite{zhou2014learning}. Two new benchmarks were added: from the 434 categories, we selected 365 categories with more than 4000 images each to create \textit{Places365-Standard} and \textit{Places365-Challenge}. 

\textbf{Places365-Standard} has 1,803,460 training images with the image number per class varying from 3,068 to 5,000. The validation set has 50 images per class and the test set has 900 images per class. Note that the experiments in this paper are reported on Places365-Standard.

\textbf{Places365-Challenge} contains the same categories as \textit{Places365-Standard}, but the training set is significantly larger with a total of ~8 million training images. The validation set and testing set are the same as the Places365-Standard. This subset was released for the Places Challenge 2016\footnote{http://places2.csail.mit.edu/challenge.html} held in conjunction with the European Conference on Computer Vision (ECCV) 2016, as part of the ILSVRC Challenge.  

\textbf{Places205}. Places205, described in  \cite{zhou2014learning}, has 2.5 million images from 205 scene categories. The image number per class varies from 5,000 to 15,000. The training set has 2,448,873 total images, with 100 images per category for the validation set and 200 images per category for the test set. 

\textbf{Places88}. Places88 contains the 88 common scene categories among the ImageNet  \cite{russakovsky2015imagenet}, SUN \cite{xiao2010sun} and Places205 databases. Note that Places88 contains only the images obtained in round 2 of annotations, from the first version of Places used in \cite{zhou2014learning}. We call the corresponding subsets Places88, ImageNet88 and SUN88. These subsets are used to compare performances across different scene-centric databases, as the three datasets contain different exemplars per category. Note that finding correspondences between the classes defined in ImageNet and Places brings some challenges. ImageNet follows the WordNet definitions, but some WordNet definitions are not always appropriate for describing places. For instance, the class 'elevator' in ImageNet refers to an object. In Places, 'elevator' takes different meanings depending on the location of the observer: elevator door, elevator interior, or elevator lobby. Many categories in ImageNet do not differentiate between indoor and outdoor (e.g., ice-skating rink) while in Places, indoor and outdoor versions are separated as they do not necessarily afford the same function. 

\subsubsection{Dataset Diversity}

Given the types of images found on the internet, some categories will be more biased than others in terms of viewpoints, types of objects, or even image style ~\cite{torralba2011unbiased}. However, bias can be compensated with a high diversity of images (with many appearances represented in the dataset). In the next section, we describe a measure of dataset diversity to compare how diverse images from three scene-centric datasets (Places88, SUN88 and ImageNet88) are.

Comparing datasets is an open problem. Even datasets covering the same visual classes have notable differences providing different generalization performances when used to train a classifier ~\cite{torralba2011unbiased}. Beyond the number of images and categories, there are aspects that are important but difficult to quantify, like the variability in camera poses, in decoration styles or in the type of objects that appear in the scene.
 
Although the quality of a database is often task dependent, it is reasonable to assume that a good database should be \textbf{dense} (with a high degree of data concentration), and \textbf{diverse} (it should include a high variability of appearances and viewpoints). Imagine, for instance, a dataset composed of 100,000 images all taken within the same bedroom. This dataset would have a very high density but a very low diversity as all the images will look very similar. An ideal dataset, expected to generalize well, should have high {\em diversity} as well.  While one can achieve high density by collecting a large number of images, diversity is not an obvious quantity to estimate in image sets, as it assumes some notion of similarity between images. One way to estimate similarity is to ask the question \emph{are these two images similar?} However, similarity in the wild is a subjective and loose concept, as two images can be viewed as similar if they contain similar objects, and/or have similar spatial configurations, and/or have similar decoration styles and so on. A way to circumvent this problem is to define {\em relative measures} of similarity for comparing datasets.

Several measures of diversity have been proposed, particularly in biology for characterizing the richness of an ecosystem (see \cite{Heip98} for a review). Here, we propose to use a measure inspired by the {\em Simpson index of diversity}~\cite{simpson1949}. The Simpson index measures the probability that two random individuals from an ecosystem belong to the same species. It is a measure of how well distributed the individuals across different species are in an ecosystem, and it is related to the entropy of the distribution. Extending this measure for evaluating the diversity of images within a category is non-trivial if there are no annotations of sub-categories. For this reason, we propose to measure the relative diversity of image datasets A and B based on the following idea: if set A is more diverse than set B, then two random images from set B are more likely to be visually similar than two random samples from A. Then, the diversity of A with respect to B can be defined as $\textrm{Div}_B (A) = 1-p( d(a_1,a_2) < d(b_1,b_2) )$, where $a_1, a_2 \in A$ and $b_1, b_2 \in B$  are randomly selected. With this definition of relative diversity we have that A is more diverse than B if, and only if, $\textrm{Div}_B (A) > \textrm{Div}_A (B)$.
For an arbitrary number of datasets, $A_1,...,A_N$:

\begin{equation}
\textrm{Div}_{A_2,...,A_N} (A_1) = 1-p(d(a_{11},a_{12}) < \min_{i=2:N} d(a_{i1},a_{i2}))
\end{equation}

where $a_{i1},a_{i2} \in A_i$ are randomly selected.

\begin{figure}
\begin{center}
\includegraphics[width=1\linewidth]{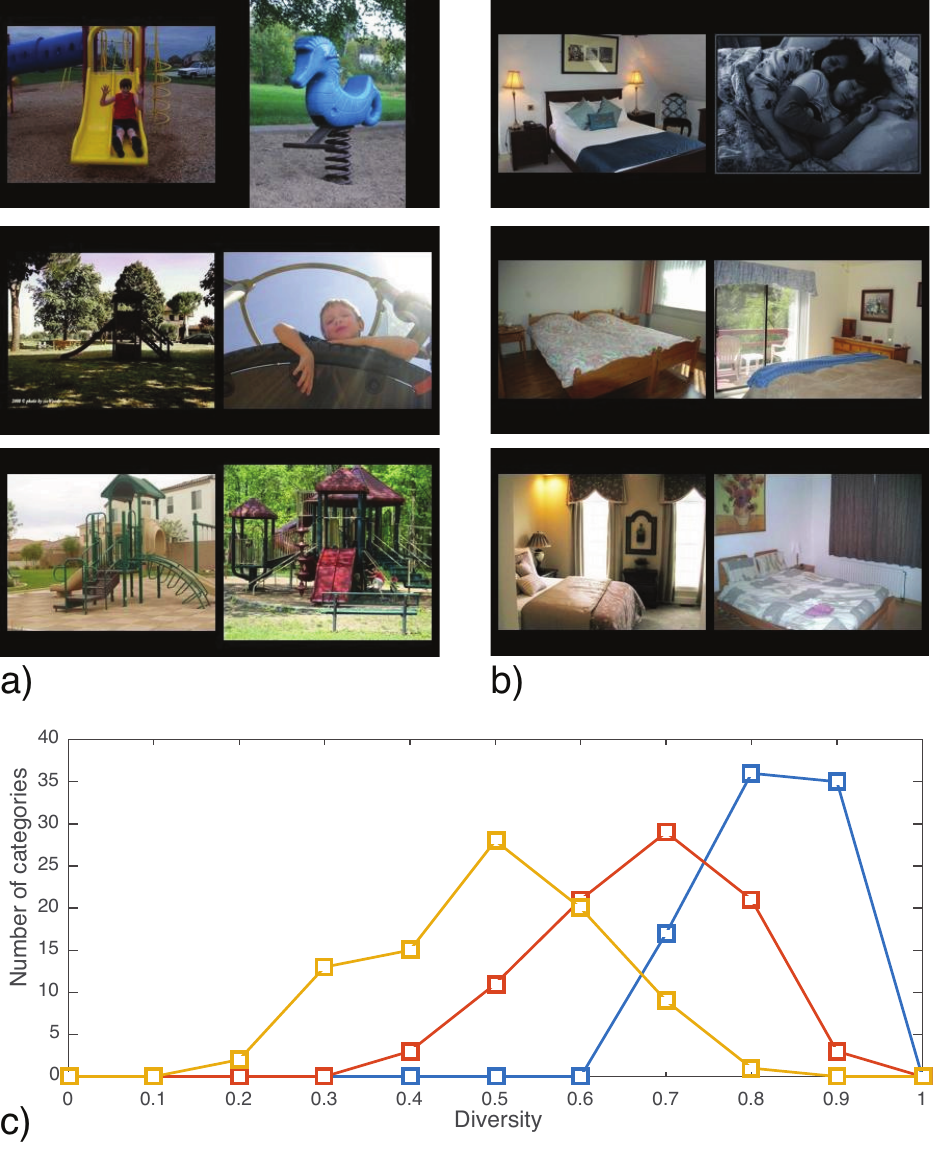}
\end{center}
 \caption{Examples of pairs for the diversity experiment for a) playground and b) bedroom. Which pair shows the most similar images? The bottom pairs were chosen in these examples. c) Histogram of relative diversity per each category (88 categories) and dataset. Places (in blue line) contains the most diverse set of images, then ImageNet (in red line) and the lowest diversity is in the SUN database (in yellow line) as most images are prototypical of their class.}
 \label{fig:divvsden}
\end{figure}

We measured the relative diversities between SUN, ImageNet and Places using AMT. Workers were presented with different pairs of images and they had to select the pair that contained the most similar images. The pairs were randomly sampled from each database. Each trial was composed of 4 pairs from each database, giving a total of 12 pairs to choose from. We used 4 pairs per database to increase the chances of finding a similar pair and avoiding users having to skip trials. AMT workers had to select the most similar pair on each trial. We ran 40 trials per category and two observers per trial, for the 88 categories in common between ImageNet, SUN and Places databases. Fig.~\ref{fig:divvsden}.a-b shows some examples of pairs from the diversity experiments for the scene categories playground (a) and bedroom (b). In the figure only one pair from each database is shown. We observed that different annotators were consistent in deciding whether a pair of images was more similar than another pair of images.

Fig.~\ref{fig:divvsden}.c shows the histograms of relative diversity for all the 88 scene categories common to the three databases. If the three datasets were identical in terms of diversity, the average diversity should be 2/3 for the three datasets. Note that this measure of diversity is a relative measure between the three datasets. In the experiment, users selected pairs from the SUN database to be the closest to each other $50\%$ of the time, while the pairs from the Places database were judged to be the most similar only on $17\%$ of the trials. ImageNet pairs were selected $33\%$ of the time. 

The results show that there is a large variation in terms of diversity among the three datasets, showing Places to be the most diverse of the three datasets. The average relative diversity on each dataset is $0.83$ for Places, $0.67$ for ImageNet and $0.50$ for SUN. To illustrate, the categories with the largest variation in diversity across the three datasets were {\em playground}, {\em veranda} and {\em waiting room}.

\section{Convolutional Neural Networks for Scene Classification}

Given the impressive performance of the deep Convolutional Neural Networks (CNNs), particularly on the ImageNet benchmark \cite{krizhevsky2012imagenet,russakovsky2015imagenet}, we choose three popular CNN architectures, \textbf{AlexNet} \cite{krizhevsky2012imagenet}, \textbf{GoogLeNet} \cite{szegedy2014going}, and \textbf{VGG} 16 convolutional-layer CNN \cite{simonyan2014very}, then train them on \textbf{Places205} and \textbf{Places365-Standard} respectively to create baseline CNN models. The trained CNNs are named as Places\textit{Subset}-\textit{CNN}, i.e., Places205-AlexNet or Places365-VGG.

All the Places-CNNs presented here were trained using the Caffe package \cite{Jia13Caffe} on Nvidia GPUs Tesla K40 and Titan X\footnote{All the Places-CNNs are available at \url{https://github.com/metalbubble/places365}}. Additionally, given the recent breakthrough performances of the Residual Network (ResNet) on ImageNet classification \cite{he2015resnet}, we further fine-tuned \textbf{ResNet152} on the Places365-Standard (termed as Places365-ResNet) and compared it with the other trained-from-scratch Places-CNNs for scene classification.

\subsection{Results on Places205 and Places365}

After training the various Places-CNNs, we used the final output layer of each network to classify the test set images of Places205 and SUN205 (see \cite{zhou2014learning}). The classification results for Top-1 accuracy and Top-5 accuracy are listed in Table \ref{CNN_test205}. As a baseline comparison, we show the results of a linear SVM trained on ImageNet-CNN features of 5000 images per category in Places205 and 50 images per category in SUN205 respectively. 

Places-CNNs perform much better than the ImageNet feature+SVM baseline while, as expected, Places205-GoogLeNet and Places205-VGG outperformed Places205-AlexNet with a large margin due to their deeper structures. To date (Oct 2, 2016) the top ranked results on the test set of Places205 leaderboard\footnote{\url{http://places.csail.mit.edu/user/leaderboard.php}} is 64.10\% on Top-1 accuracy and 90.65\% on Top-5 accuracy. Note that for the test set of SUN205, we didn't fine-tune the Places-CNNs on the training set of SUN205, as we directly evaluated them on the test set of SUN.

\begin{table*}\caption{Classification accuracy on the test set of Places205 and the test set of SUN205. We use the class score averaged over 10-crops of each test image to classify the image. $*$ shows the top 2 ranked results from the Places205 leaderboard.}\label{CNN_test205}
\begin{center}
\begin{tabular}{lccccc}
\hline
\hline
& \multicolumn{2}{c}{Test set of Places205}& & \multicolumn{2}{c}{Test set of SUN205}\\
\cline{2-3} \cline{5-6}
& Top-1 acc. & Top-5 acc. & &Top-1 acc. & Top-5 acc. \\
\hline
ImageNet-AlexNet feature+SVM & 40.80\% & 70.20\% && 49.60\%& 80.10\% \\
Places205-AlexNet & 50.04\% & 81.10\%  & & 67.52\% & 92.61\% \\
Places205-GoogLeNet & 55.50\% & 85.66\% &  & 71.6\% & 95.01\% \\
Places205-VGG & \textbf{58.90}\% & \textbf{87.70}\% & & \textbf{74.6}\% & \textbf{95.92}\% \\
\hline
SamExynos$*$  & 64.10\% & 90.65\% &  & - & - \\
SIAT MMLAB$*$ & 62.34\% & 89.66\% &  & - & - \\
\hline \\
\end{tabular}
\end{center}
\end{table*}

We further evaluated the baseline Places365-CNNs on the validation set and test set of Places365 shown in Fig.\ref{CNN_test365}.  Places365-VGG and Places365-ResNet have similar top performances compared with the other two CNNs\footnote{The performance of the ResNet might result from fine-tuning or under-training, as the ResNet is not trained from scratch.}. 
Even if Places365 has 160 more categories than Places205, the Top-5 accuracy of the Places205-CNNs (trained on the previous version of Places \cite{zhou2014learning}) on the test set only drops by ~2.5\%.

\begin{figure*}
\begin{center}
\includegraphics[width=1\linewidth]{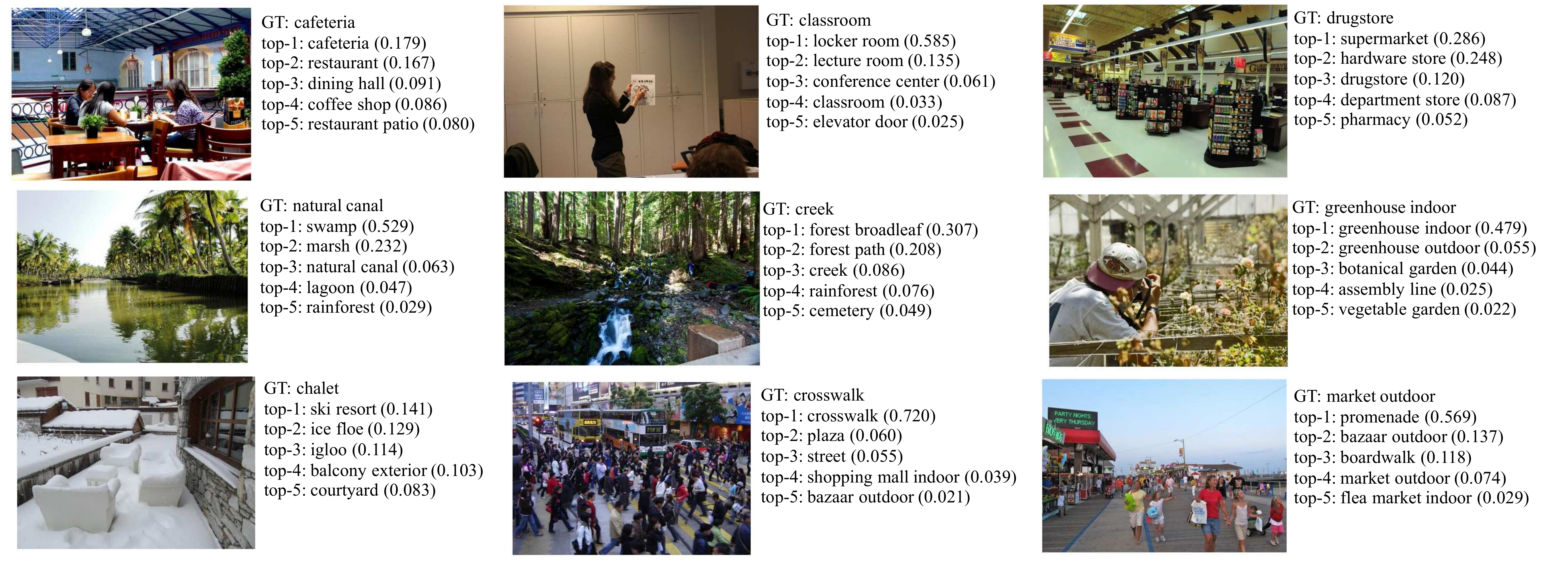}
\end{center}
 \caption{The predictions given by the Places365-VGG for the images from the validation set. The ground-truth label (GT) and the top 5 predictions are shown. The number beside each label indicates the prediction confidence.}
 \label{fig:qualitativeresult}
\end{figure*}

\begin{figure}
\begin{center}
\includegraphics[width=1\linewidth]{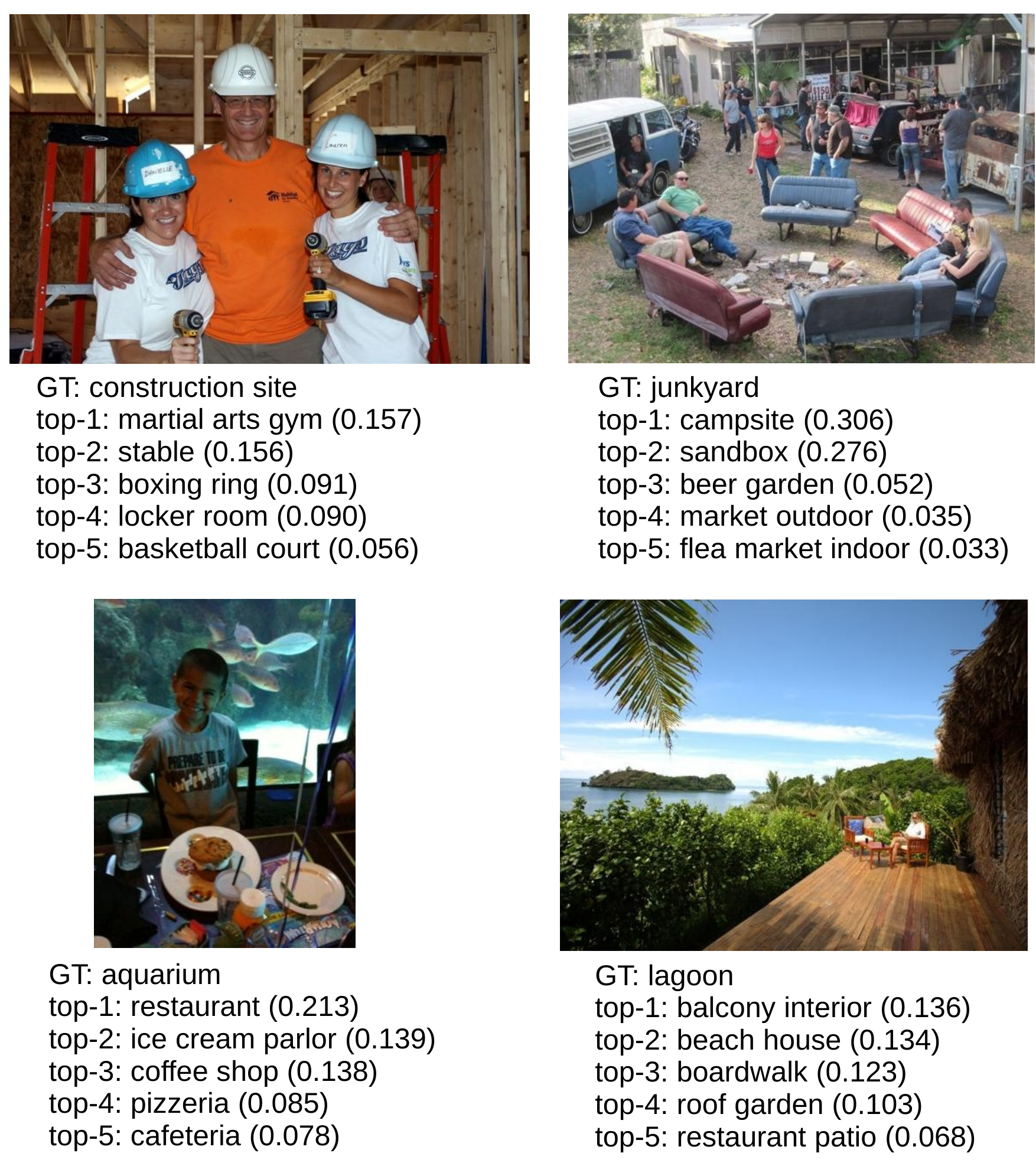}
\end{center}
\caption{Examples of predictions rated as incorrect in the validation set by the Places365-VGG. GT states for ground truth label. Note that some of the top-5 responses are often not wrong per se, predicting semantic categories near by the GT category. See the text for details.}
   \label{fig:wrongprediction}
\end{figure}

Fig.\ref{fig:qualitativeresult} shows the responses to examples correctly predicted by the Places365-VGG. Most of the Top-5 responses are very relevant to the scene description. Some failure or ambiguous cases are shown in Fig.\ref{fig:wrongprediction}: Broadly, we can identify two kinds of misclassification given the current label attribution of Places: 1) less-typical activities happening in a scene, such as taking group photo in a construction site and camping in a junkyard; 2) images composed of multiple scene parts, which make one ground-truth scene label not sufficient to describe the whole environment. These illustrate the need to have multi-ground truth labels for describing environments. 

It is important to emphasize that for many scene categories the Top-1 accuracy might be an ill-defined measure: environments are inherently multi-labels in terms of their semantic description. Different observers will use different terms to refer to the same place, or different parts of the same environment, and all the labels might fit well the description of the scene. This is obvious in the examples of Fig.\ref{fig:wrongprediction}. Future development of the Places database, and the Places Challenge, will explore to assign multiple ground truth labels or free-form sentences to images to better capture the richness of visual descriptions inherent to environments. 

\begin{table*}\caption{Classification accuracy on the validation set and test set of Places365. We use the class score averaged over 10-crops of each testing image to classify the image.}\label{CNN_test}
\centering
\begin{tabular}{lccccc}
\hline
\hline
& \multicolumn{2}{c}{Validation Set of Places365}& & \multicolumn{2}{c}{Test Set of Places365}\\
\cline{2-3} \cline{5-6}
& Top-1 acc. & Top-5 acc. & &Top-1 acc. & Top-5 acc. \\
\hline
Places365-AlexNet & 53.17\% & 82.89\%  & & 53.31\% & 82.75\% \\
Places365-GoogLeNet & 53.63\% & 83.88\% &  & 53.59\% & 84.01\% \\
Places365-VGG & \textbf{55.24}\% & 84.91\% & & \textbf{55.19}\% & 85.01\% \\
Places365-ResNet & 54.74\% & \textbf{85.08}\% & & 54.65\% & \textbf{85.07}\% \\
\hline
\end{tabular}
\label{CNN_test365}
\end{table*}

\subsection{Web-demo for Scene Recognition}

\begin{figure}
\begin{center}
\includegraphics[width=1\linewidth]{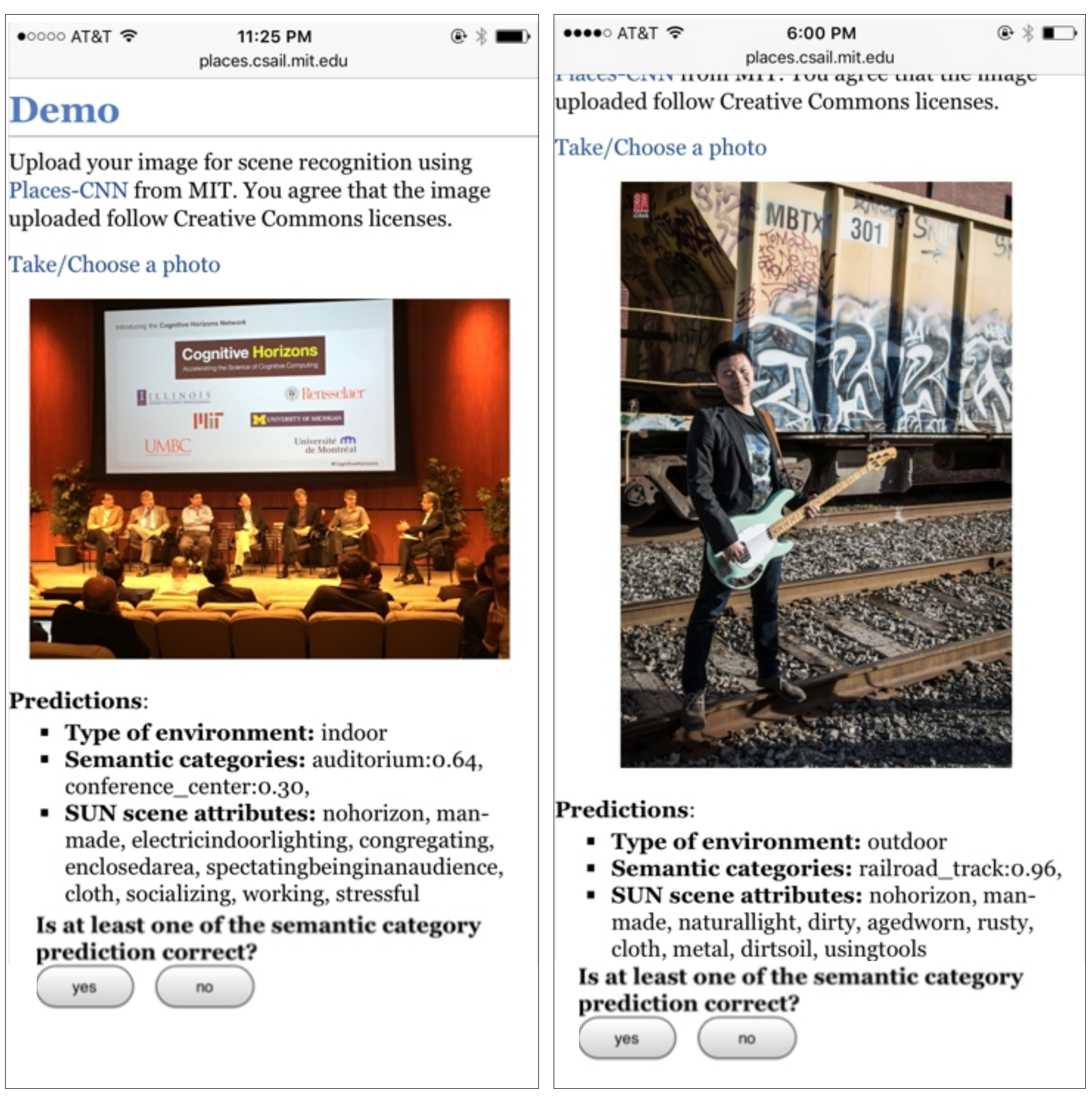}
\end{center}
 \caption{Two screenshots of the scene recognition demo based on the Places-CNN. The web-demo predicts the type of environment, the semantic categories, and associated scene attributes for uploaded photos. }
 \label{fig:webdemo}
\end{figure}

Based on the Places-CNN we trained, we created a web-demo for scene recognition\footnote{\url{http://places.csail.mit.edu/demo.html}}, accessible through a computer browser or mobile phone. People can upload photos to the web-demo to predict the type of environment, with the 5 most likely semantic categories, and relevant scene attributes. Two screenshots of the prediction result on the mobile phone are shown in Fig.\ref{fig:webdemo}. Note that people can submit feedback about the result. The top-5 recognition accuracy of our recognition web-demo in the wild is about 72\% (from the 9,925 anonymous feedbacks dated from Oct.19, 2014 to May 5, 2016), which is impressive given that people uploaded all kinds of photos from real-life and not necessarily places-like photos (these results are for Places205-AlexNet as the back-end prediction model in the demo).

\subsection{Generic Visual Features from ImageNet-CNNs and Places-CNNs}

We further used the activation from the trained Places-CNNs as generic features for visual recognition tasks using different image classification benchmarks. Activations from the higher-level layers of a CNN, also termed \textit{deep features}, have proven to be effective generic features with state-of-the-art performance on various image datasets \cite{donahue2014decaf,razavian2014cnn}. But most of the deep features are from the CNNs trained on ImageNet, which is mostly an object-centric dataset. 

Here we evaluated the classification performances of the deep features from object-centric CNNs and scene-centric CNNs in a systematic way. The deep features from several Places-CNNs and ImageNet-CNNs on the following scene and object benchmarks are tested: SUN397 \cite{xiao2010sun}, MIT Indoor67 \cite{quattoni2009recognizing}, Scene15 \cite{lazebnik2006beyond}, SUN Attribute \cite{patterson2012sun}, Caltech101 \cite{fei2007learning}, Caltech256 \cite{griffin2007caltech}, Stanford Action40 \cite{yao2011human}, and UIUC Event8 \cite{li2007and}. 

All of the  experiments follow the standards in those papers. In the SUN397 experiment \cite{xiao2010sun}, the training set size is 50 images per category. Experiments were run on 5 splits of the training set and test set given in the dataset. In the MIT Indoor67 experiment \cite{quattoni2009recognizing}, the training set size is 100 images per category. The experiment is run on the split of the training set and test set given in the dataset. In the Scene15 experiment \cite{lazebnik2006beyond}, the training set size is 50 images per category. Experiments are run on 10 random splits of the training set and test set. In the SUN Attribute experiment \cite{patterson2012sun}, the training set size is 150 images per attribute. The reported result is the average precision. The splits of the training set and test set are given in the paper. In Caltech101 and Caltech256 experiment \cite{fei2007learning,griffin2007caltech}, the training set size is 30 images per category. The experiments are run on 10 random splits of the training set and test set. In the Stanford Action40 experiment \cite{yao2011human}, the training set size is 100 images per category. Experiments are run on 10 random splits of the training set and test set. The reported result is the classification accuracy. In the UIUC Event8 experiment \cite{li2007and}, the training set size is 70 images per category and the test set size is 60 images per category. The experiments are run on 10 random splits of the training set and test set.

Places-CNNs and ImageNet-CNNs have the same network architectures for AlexNet, GoogLeNet, and VGG, but they are trained on scene-centric data and object-centric data respectively. For AlexNet and VGG, we used the 4096-dimensional feature vector from the activation of the Fully Connected Layer (\texttt{fc7}) of the CNN. For GoogLeNet, we used the 1024-dimensional feature vector from the response of the global average pooling layer before softmax producing the class predictions. The classifier in all of the experiments is a linear SVM with the default parameter for all of the features. 
 
\begin{table*}\caption{Classification accuracy/precision on scene-centric databases (the first four datasets) and object-centric databases (the last four datasets) for the deep features of various Places-CNNs and ImageNet-CNNs. All the accuracy/precision is the top-1 accuracy/precision.}\label{dataset_comparison}
\centering
\footnotesize
\begin{tabular}{l|cccc|cccc|c}
%\begin{tabular}{lllllllll}
\hline
\hline
Deep Feature & SUN397 & MIT Indoor67 & Scene15 & SUN Attribute & Caltech101&Caltech256&Action40 & Event8 & Average \\
\hline
Places365-AlexNet  & \textbf{56.12} & \textbf{70.72}  & 89.25 & 92.98 & 66.40 & 46.45 & 46.82 & 90.63 & 69.92 \\
Places205-AlexNet  & 54.32 & 68.24  & \textbf{89.87} & \textbf{92.71} & 65.34 & 45.30 & 43.26 & 94.17 & 69.15\\
ImageNet-AlexNet   & 42.61 & 56.79  & 84.05 & 91.27 & \textbf{87.73} & \textbf{66.95} & \textbf{55.00} & \textbf{93.71} & 72.26\\
\hline
Places365-GoogLeNet & \textbf{58.37} & 73.30 & \textbf{91.25}& \textbf{92.64} & 61.85 & 44.52 & 47.52 & 91.00 & 70.06 \\
Places205-GoogLeNet & 57.00 & \textbf{75.14} & 90.92& 92.09 & 54.41 & 39.27 & 45.17 & 92.75 & 68.34\\
ImageNet-GoogLeNet  & 43.88 & 59.48 & 84.95& 90.70 & \textbf{89.96} & \textbf{75.20} & \textbf{65.39} & \textbf{96.13} & 75.71 \\
\hline
Places365-VGG & \textbf{63.24} & 76.53 & \textbf{91.97} & \textbf{92.99} & 67.63 & 49.20 & 52.90 & 90.96 & 73.18\\
Places205-VGG & 61.99 & \textbf{79.76} & 91.61 & 92.07 & 67.58 & 49.28 & 53.33 & 93.33 & 73.62\\
ImageNet-VGG  & 48.29 & 64.87 & 86.28 & 91.78 & \textbf{88.42} & \textbf{74.96} & \textbf{66.63} & \textbf{95.17} & 77.05\\
\hline
%Places365-ResNet152 & - & - & - & - & - & - & - & - & - \\
%ImageNet-ResNet152 & 55.16 & 73.56 & 90.35 & 92.79 & 92.11 & 81.58 & 74.00 & 96.29 & - \\
\hline
Hybrid1365-VGG  & 61.77 & 79.49 & 92.15 & 92.93 & 88.22 & 76.04 & 68.11 & 93.13 & \textbf{81.48}\\
%Hybrid1365-ResNet152 & - & - & - & - & - & - & - & - & - \\
\hline
%Hybrid-AlexNet   \\
%Hybrid-VGG  \\
%\hline
% &Caltech101&Caltech256&Action40 & Event8 \\
% \hline
% Places401-AlexNet & 68.20 & 47.92 & 48.71$\pm$0.29 & 94.62$\pm$1.00 \\
% Places201-AlexNet & 65.18$\pm$0.88 & 45.59$\pm$0.31 &42.86$\pm$0.25 & 94.12$\pm$0.99 \\
% ImageNet-AlexNet & 87.22$\pm$0.92 & 67.23$\pm$0.27 & 54.92$\pm$0.33 & 94.42$\pm$0.76 \\
% \hline
% Places401-GoogLeNet & & & & \\
% Places201-GoogLeNet & & & & \\
% ImageNet-GoogLeNet & & & & \\
% \hline
% Places401-VGG & & & & \\
% Places201-VGG & & & & \\
% ImageNet-VGG & & & & \\
\end{tabular}
\label{tableResultsDeepSceneFeat}
\end{table*}

% \begin{table}\caption{Avereage accuracy on scene-centric datasets and object-centric datasets and their average.}\label{dataset_comparison}
% \centering
% \footnotesize
% \begin{tabular}{lcccccccc}
% %\begin{tabular}{lllllllll}
% \hline
% \hline
% Deep Feature & scene-centric & object-centric & all \\
% \hline
% Places365-AlexNet  & 77.27 & 62.58 & 69.92\\
% Places205-AlexNet  & 76.29 & 62.02 & 69.15\\
% Places365-GoogLeNet & 78.89 & 61.22 & 70.06\\
% Places205-GoogLeNet & 78.79 & 57.90 & 68.34\\
% Places365-VGG & 81.18 & 65.17 & 73.18\\
% Places205-VGG & 81.36 & 65.88 & 73.62\\
% ImageNet-AlexNet   & 68.68 & 75.85 & 72.26\\
% ImageNet-GoogLeNet  & 69.75 & 81.67 & 75.71\\
% ImageNet-VGG & 72.81 & 81.29 & 77.05\\ 
% Hybrid1365-VGG & 81.31 & 81.23 & 81.27\\
% \hline
% \end{tabular}
% \label{tableResults_mean}
% \end{table}

\begin{figure}
\begin{center}
\includegraphics[width=1\linewidth]{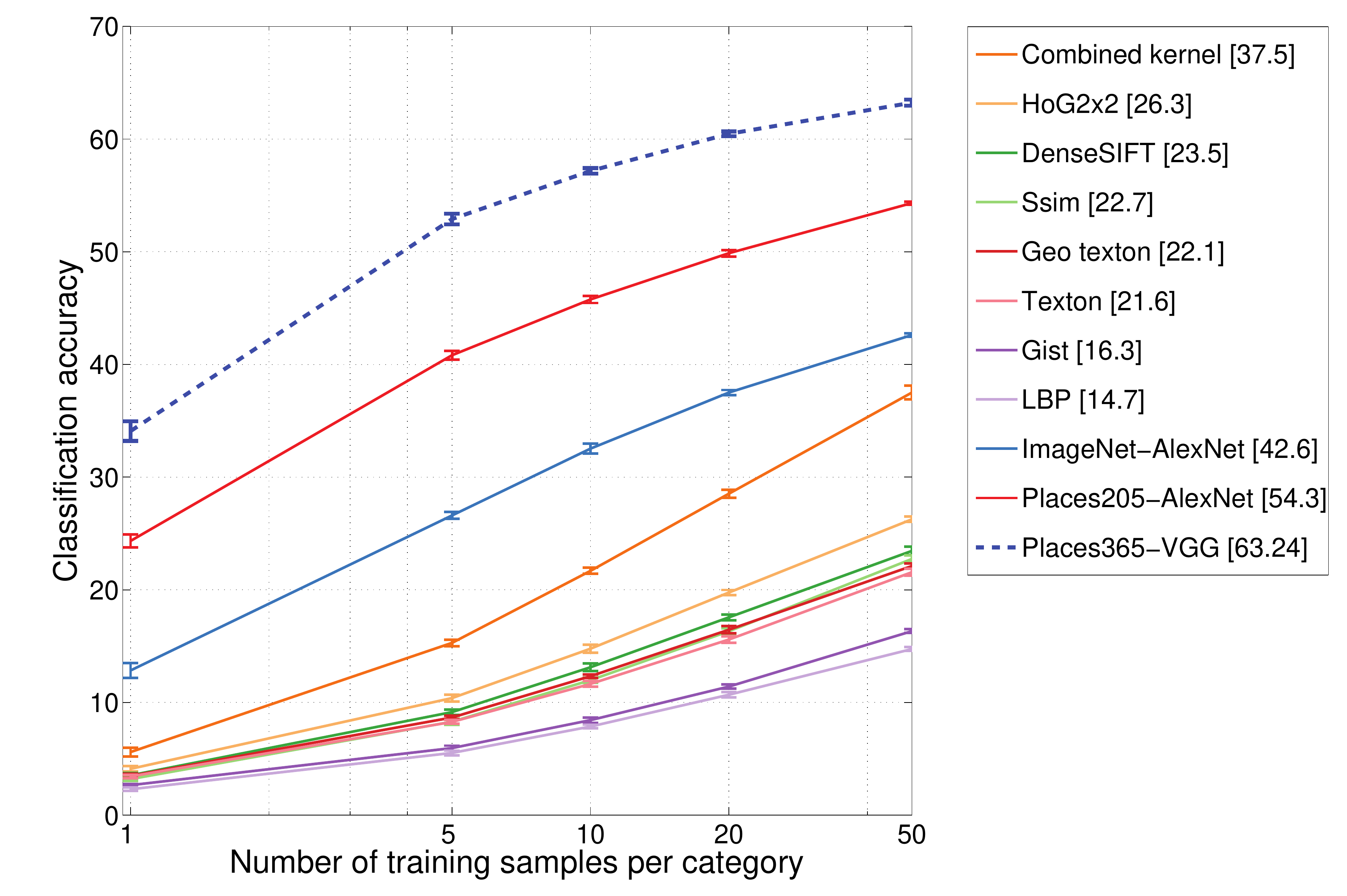}
\end{center}
  \caption{Classification accuracy on the SUN397 Dataset. We compare the deep features of Places365-VGG, Places205-AlexNet (result reported in \cite{zhou2014learning}), and ImageNet-AlexNet, to those hand-designed features. The deep features of Places365-VGG outperforms other deep features and hand-designed features in large margins. Results of other hand-designed features/kernels are fetched from \cite{xiao2010sun}.}
  \label{SUN_SUNattribute}
\end{figure}

Table \ref{tableResultsDeepSceneFeat} summarizes the classification accuracy on various datasets for the deep features of Places-CNNs and the deep features of the ImageNet-CNNs. Fig.\ref{SUN_SUNattribute} plots the classification accuracy for different visual features on the SUN397 database over different numbers of training samples per category. The classifier is a linear SVM with the same default parameters for the two deep feature layers (C=1) \cite{linearSVM}. The Places-CNN features show impressive performance on scene-related datasets, outperforming the ImageNet-CNN features. On the other hand, the ImageNet-CNN features show better performance on object-related image datasets. Importantly, our comparison shows that Places-CNN and ImageNet-CNN have complementary strengths on scene-centric tasks and object-centric tasks, as expected from the type of the datasets used to train these networks. On the other hand, the deep features from the Places365-VGG achieve the best performance (63.24\%) on the most challenging scene classification dataset SUN397, while the deep features of Places205-VGG performs the best on the MIT Indoor67 dataset. As far as we know, they are the state-of-the-art scores achieved by a single feature + linear SVM on those two datasets. Furthermore, we merge the 1000 classes from the ImageNet and the 365 classes from the Places365-Standard to train a VGG (Hybrid1365-VGG). The deep feature from the Hybrid1365-VGG achieves the best score averaged over all the eight image datasets.

\subsection{Visualization of the Internal Units and the CNNs}

\begin{figure*}
\begin{center}
\includegraphics[width=1\linewidth]{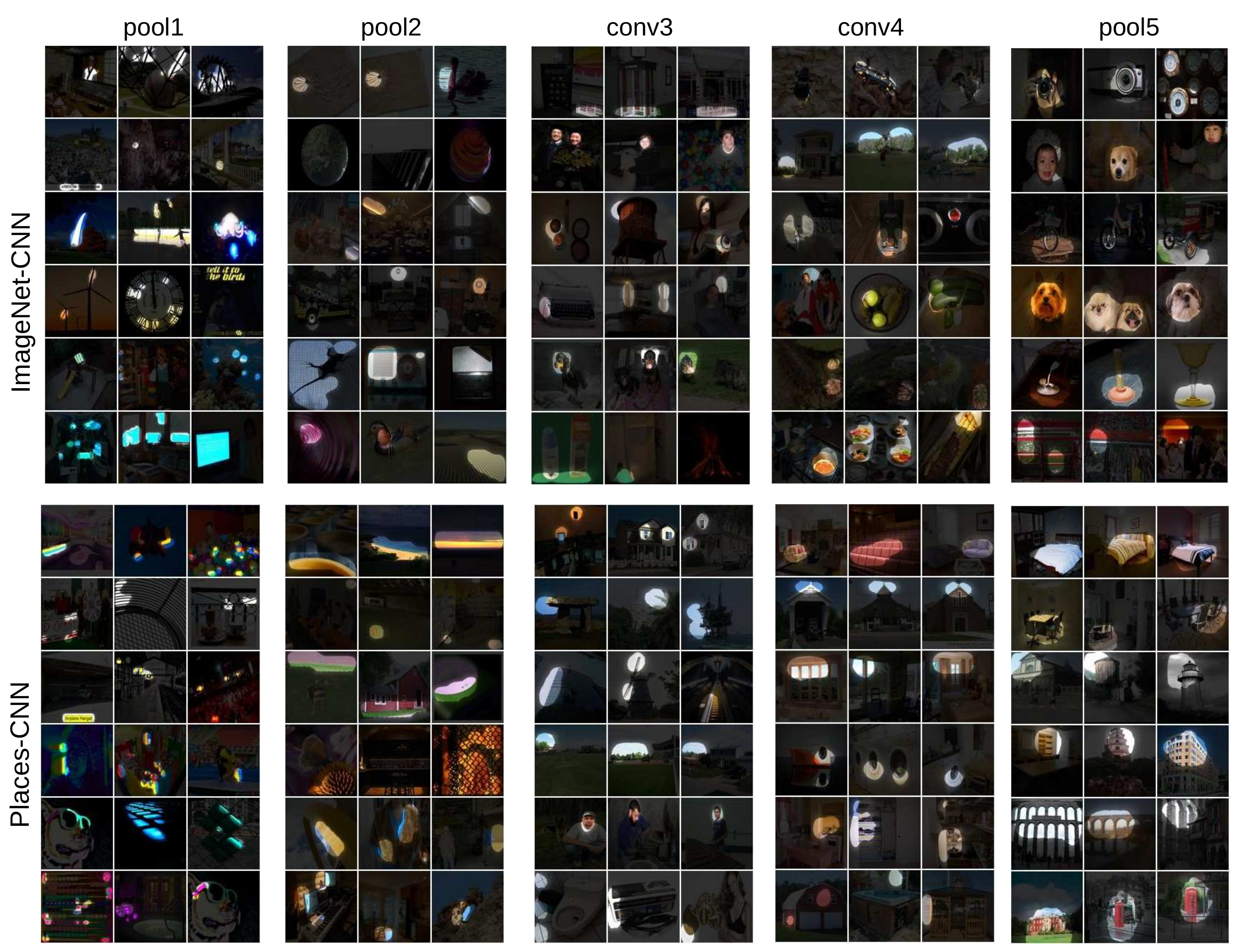}
\end{center}
\caption{a) Visualization of the units' receptive fields at different layers for the ImageNet-CNN and Places-CNN. Subsets of units at each layer are shown. In each row we show the top 3 most activated images. Images are segmented based on the estimated receptive fields of the units at different layers of ImageNet-CNN and Places-CNN. Here we take ImageNet-AlexNet and Places205-AlexNet as the comparison examples. See the detailed visualization methodology in \cite{zhou2014object}.}
   \label{visualization}
\end{figure*}

Through the visualization of the units responses for various levels of network layers, we can have a better understanding of what has been learned inside CNNs and what are the differences between the object-centric CNN trained on ImageNet and the scene-centric CNN trained on Places given that they share the same architecture (here we use AlexNet). Following the methodology in \cite{zhou2014object}， we estimated the receptive fields of the units in the Places-CNN and ImageNet-CNN. Then we segmented the images with high unit activation using the estimated receptive fields. The image segmentation results by the receptive fields of units from different layers are shown in Fig.\ref{visualization}. We can see that from \texttt{pool1} to \texttt{pool5}, the units detect visual concepts from low-level edge/texture to high-level object/scene parts. Furthermore, in the object-centric ImageNet-CNN there are more units detecting object parts such as dog and people's heads in the \texttt{pool5} layer, while in the scene centric Places-CNN there are more units detecting scene parts such as bed, chair, or buildings in the \texttt{pool5} layer.

Thus the specialty of the units in the object-centric CNN and scene-centric CNN yield very different performances of generic visual features on a variety of recognition benchmarks (object-centric datasets vs scene-centric datasets) in Table \ref{tableResultsDeepSceneFeat}. 

\begin{figure*}
\begin{center}
\includegraphics[width=1\linewidth]{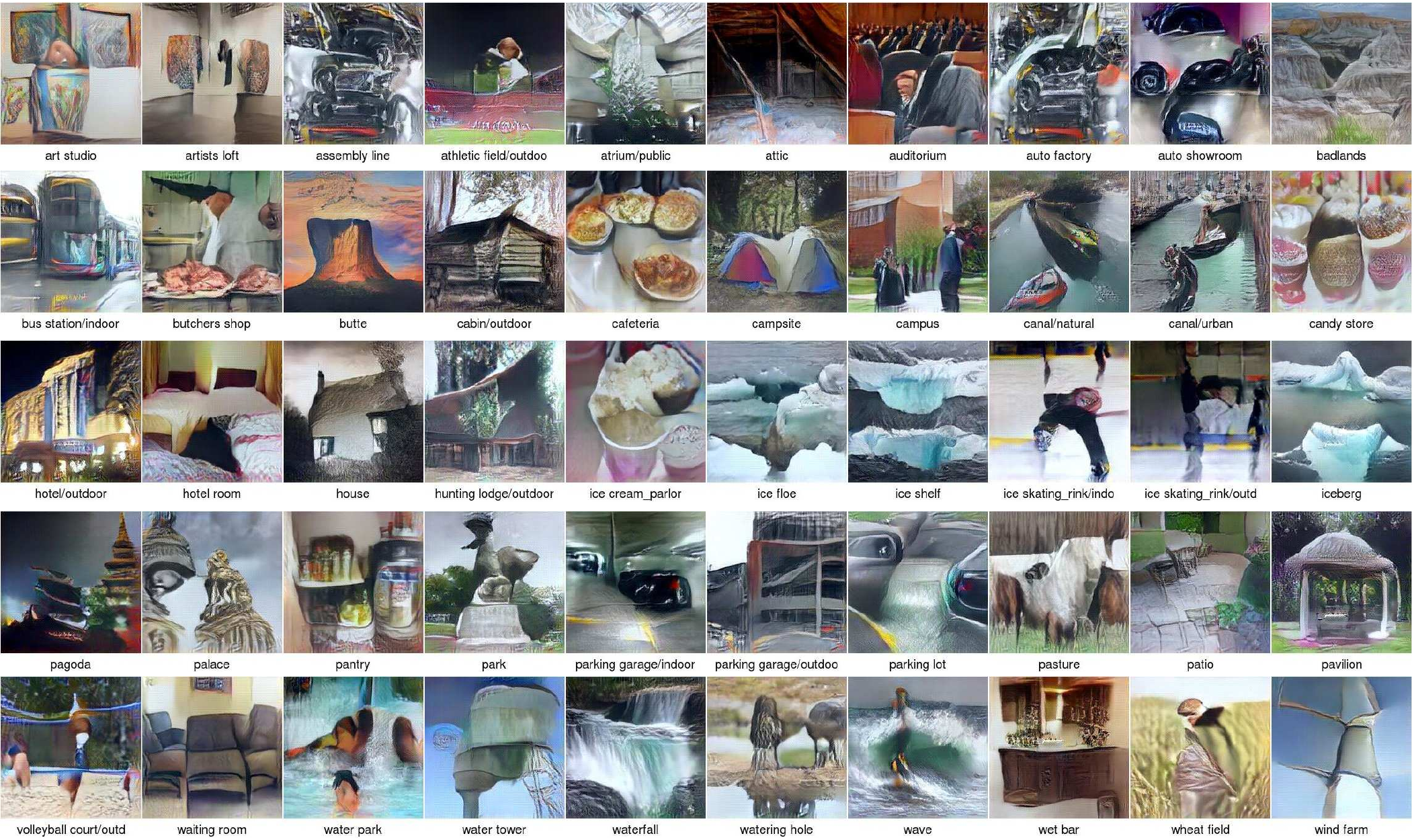}
\end{center}
\caption{The synthesized images preferred by the final output of Places365-AlexNet for 50 scene categories.}
   \label{category_synthetic}
\end{figure*}

\begin{figure*}
\begin{center}
\includegraphics[width=1\linewidth]{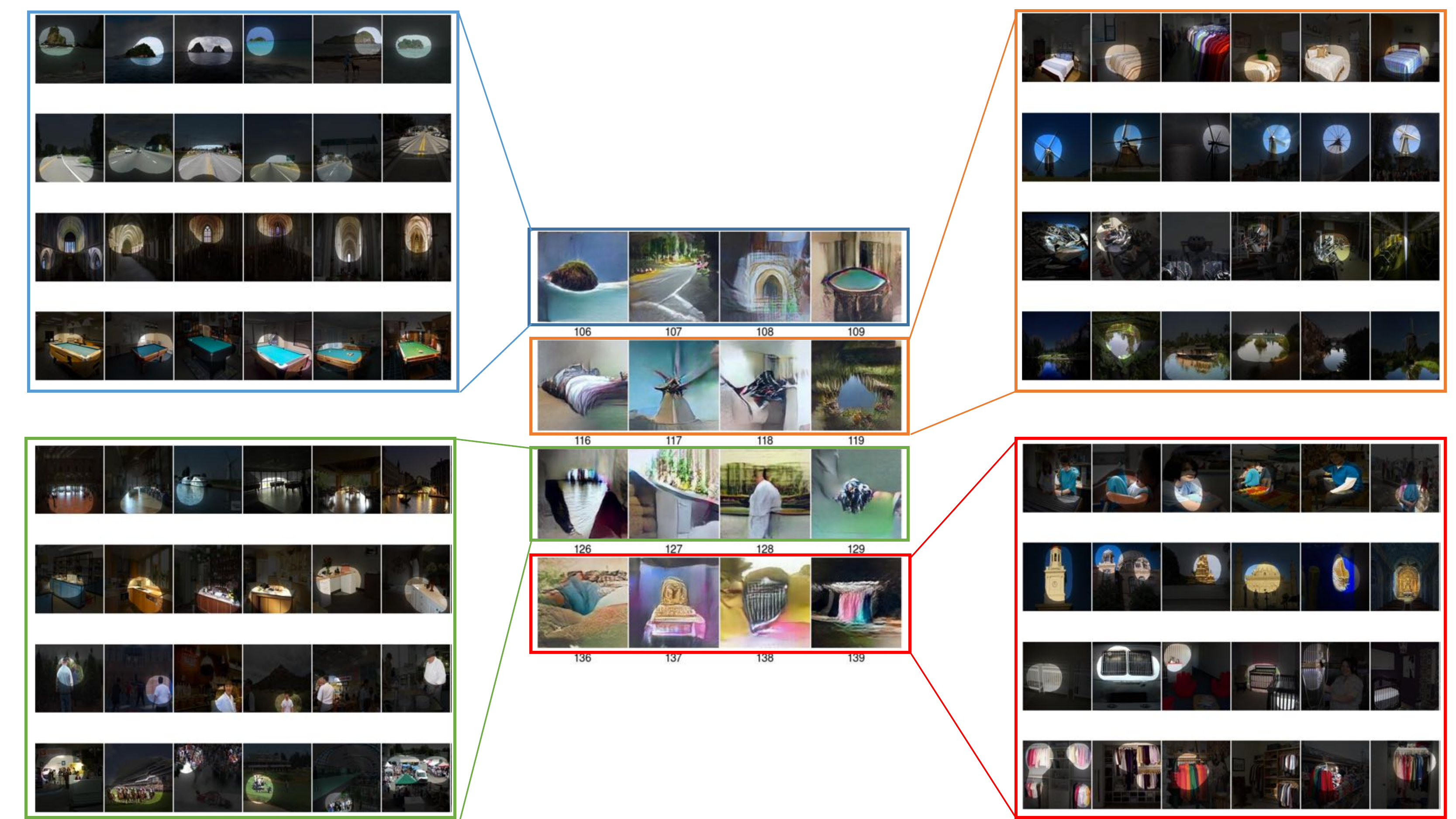}
\end{center}
\caption{The synthesized images preferred by the \texttt{pool5} units of the Places365-AlexNet corresponds to the segmented images by the receptive fields of those units. The synthetic images are very similar to the segmented image regions of the units. Each row of the segmented images correspond to one unit.}
   \label{unit_synthetic}
\end{figure*}

We further synthesized preferred input images for the Places-CNN by using the image synthesis technique proposed in \cite{nguyen2016synthesizing}. This method uses a learned prior deep generator network to generate images which maximize the final class activation or the intermediate unit activation of the Places-CNN. The synthetic images for 50 scene categories are shown in Fig.\ref{category_synthetic}. These abstract image contents reveal the knowledge of the specific scene learned and memorized by the Places-CNN: examples include the buses within a road environment in the bus station, and the tents surrounded by forest-types of features for the campsite. Here we used Places365-AlexNet (other Places365-CNNs generated similar results). We further used the synthesis technique to generate the images preferred by the units in the \texttt{pool5} layer of Places365-AlexNet. As shown in Fig.\ref{unit_synthetic}, the synthesized images are very similar to the segmented image regions by the estimated receptive field of the units.

%\subsubsection{Dark knowledge}

%G. E. Hinton, O. Vinyals, and J. Dean. Distilling the knowledge in a neural network. In NIPS 2014 Deep Learning workshop

%C. Bucilua, R. Caruana, and A. Niculescu-Mizil. Model compression. In ACM SIGKDD, 2006.

\section{Conclusion}

From the Tiny Image dataset \cite{torralba200880}, to ImageNet  \cite{deng2009imagenet} and Places  \cite{zhou2014learning},  the rise of multi-million-item dataset initiatives and other densely labeled datasets \cite{lin2014microsoft,zhou2016semantic,everingham2010pascal,cordts2016cityscapes} have enabled data-hungry machine learning algorithms to reach near-human semantic classification of visual patterns, like objects and scenes. With its high-coverage and high-diversity of exemplars, Places offers an ecosystem of visual context to guide progress on currently intractable visual recognition problems. Such problems could include determining the actions happening in a given environment, spotting inconsistent objects or human behaviors for a particular place, and predicting future events or the cause of events given a scene.

\section*{Acknowledgments}
The authors would like to thank Santani Teng, Zoya Bylinskii, Mathew Monfort and Caitlin Mullin for comments on the paper. Over the years, the Places project was supported by the National Science Foundation under Grants No. 1016862 to A.O and No. 1524817 to A.T; ONR N000141613116 to A.O; as well as MIT Big Data Initiative at CSAIL, Toyota, Google, Xerox and Amazon Awards, and a hardware donation from NVIDIA Corporation, to A.O and A.T. B.Z is supported by a Facebook Fellowship.

\ifCLASSOPTIONcaptionsoff
  \newpage
\fi

\bibliographystyle{IEEEtran}
\bibliography{egbib.bib}

% \begin{IEEEbiography}[]{Bolei Zhou}
% \end{IEEEbiography}
% \vspace{-8mm}
% \begin{IEEEbiography}[]{Agata Lapedriza}
% \end{IEEEbiography}
% \vspace{-8mm}
% \begin{IEEEbiography}[]{Jianxiong Xiao}

% \end{IEEEbiography}
% \vspace{-8mm}
% \begin{IEEEbiography}[]{Antonio Torralba)}
% \vspace{-8mm}
% \end{IEEEbiography}

% \begin{IEEEbiography}[]{Aude Oliva)}
% \end{IEEEbiography}

\end{document}